# Sentiment Classification of Food Reviews

**Hua Feng**
Department of Electrical Engineering
Stanford University
Stanford, CA 94305
fengh15@stanford.edu

**Ruixi Lin**
Department of Electrical Engineering
Stanford University
rlin2@stanford.edu

## Abstract

Sentiment analysis of reviews is a popular task in natural language processing. In this work, the goal is to predict the score of food reviews on a scale of 1 to 5 with two recurrent neural networks that are carefully tuned. As for baseline, we train a simple RNN for classification. Then we extend the baseline to modified RNN and GRU. In addition, we present two different methods to deal with highly skewed data, which is a common problem for reviews. Models are evaluated using accuracies.

## 1    Introduction

Binary classification of sentiment on reviews are an increasingly popular task in NLP. Instead of classifying positive reviews and negative reviews, we classify reviews into extremely negative, negative, neutral, positive, and extremely positive classes directly from the reviewer's score on a topic. We train a simple RNN classifier, a modified RNN classifier and a GRU classifier. Our analysis could be a useful tool to help restaurants better understand reviewers' sentiment about food, and can be used for other tasks such as recommender systems.

## 2    Problem Statement

In order to predict sentiments of reviews, we label each review with a reviewer's score indicating the sentiment of the reviewer. Our task is to predict a reviewer's score on a scale of 1 to 5, where 1 indicates the reviewer extremely dislikes the food he or she mentions in the review and 5 indicates the user likes the food a lot.

## 3    Related Work

Traditional approaches on sentiment analysis use word count or frequencies in the text which are assigned sentiment value by expert[1]. These approaches disregard the order of words. A recurrent neural network (RNN)[2] can be used for sequence labeling on sequential data of variable length, which is natural for sentiment analysis tasks where the input sentence is viewed as a sequence of tokens. Recent works explore the Gated Recurrent Units neural

network(GRU)[3] on the task of sentiment classification. GRUs are a special case of the Long Short-Term(LSTM) neural network architecture. GRUs are effective in this task because of their ability to remember long time dependencies. Furthermore, GRUs are faster to train and converge than LSTM networks.

## 4     Dataset

We work on the Amazon Fine Food Reviews dataset[4] which contains 568,454 reviews. The dataset consists of a single CSV file, which includes the ids of the products, ids of the reviewers, the scores(rating between 1 and 5) given by the reviewers, the timestamp for each review, a brief summary for each review, and the text of the reviews. We extract the columns of scores and review texts as our labels and raw inputs. Sample reviews with different scores are shown below:

| Review | Score |
|---|---|
| Product arrived labeled as Jumbo Salted Peanuts... the peanuts were actually small sized unsalted. Not sure if this was an error or if the vendor intended to represent the product as "Jumbo". | 1 |
| I have bought several of the Vitality canned dog food products and have found them all to be of good quality. The product looks more like a stew than a processed meat and it smells better. My Labrador is finicky and she appreciates this product better than most. | 5 |

In order to perform mini-batch training for the neural network models, we want tokens within each slice of epoch to come from the same review. To make this happen, we need to compensate reviews with <unk>s to the maximum length of all reviews. To introduce as fewer <unk>s as possible, we do not want the reviews differ greatly in length. In this case, we would like to keep only reviews of similar lengths. We need to determine the range of lengths of reviews. In our analysis of the original dataset, we found that the average length of reviews is 80, so we choose reviews between 75 and 87 tokens and generate a dataset of 34,091 reviews.

Another problem with the dataset is that the reviews are skewed towards higher scores, especially towards the highest score, which is 5. In the 34,091 reviews, 3,550 reviews are labeled with 1, 2,085 reviews are labeled with 2, 2,844 are labeled with 3, while 4,971 reviews are labeled with 4 and an even larger volume of 20,641 reviews are labeled with 5. As is shown in figure 1, score-2 class has the lowest number of reviews, which may lead to difficulty in predicting score-2. Score-5 class has the highest number of reviews as expected, which is around ten times of that of score-2 class. To take care of the skewedness issue, we introduce two resampling methods to produce a more balanced dataset. The methods will be discussed in section 6.

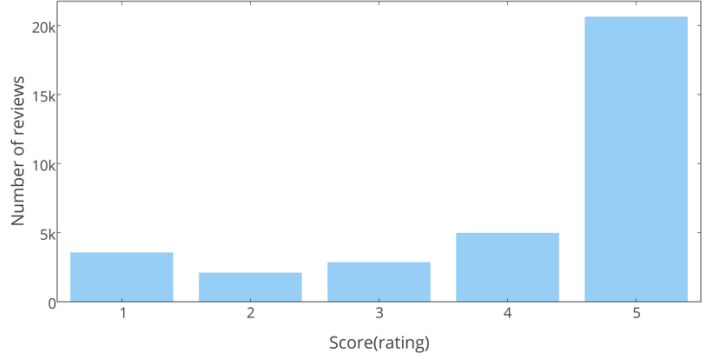

Figure 1: Number of reviews of each score in the Amazon Food Reviews dataset.

# 5 Mathematical Formulations

## 5.1 Modified Recurrent Neural Network(RNN)

Our version of RNN is a slightly modified version of the standard RNN. Instead of providing classification prediction at each word, we build the model to output prediction at the end of each epoch slice. We make this modification in order to reduce the influence of frequent words on the prediction and backpropagation.

Let T represents the number of steps, For each epoch slice $x^{(t)}, \ldots, x^{(t+T-1)}$, the forward propagation is defined as:

$$h^{(t+k)} = \sigma\big(W^{(hh)}h^{(t+k-1)} + W^{(hx)}x^{(t+k)} + b_1\big) \tag{1}$$

$$\hat{y}^{(t+T-1)/T} = softmax(W^{(s)}h^{(t+T-1)} + b_2) \tag{2}$$

Where k = 0, 1,…T-1, $x^{(t+k)}$ is the word vector embedding for the (t+k) th word in the review, $h^{(t+k)}$ is the (t+k)th hidden layer and $\hat{y}^{(t+T-1)/T}$ is the prediction output at the (t+T-1)/T th epoch slice. Details of implementation can be seen in section 6.2.

Cross-entropy error is used as loss function, the expression for a corpus size of K is as follow:

$$J = -\frac{T}{K}\sum_{t=1}^{T/K} J^{(Kt)}(\theta) = -\frac{T}{K}\sum_{t=1}^{T/K}\sum_{c=1}^{C} y_{t,c} \log(y_{t,c}) \tag{3}$$

Where T is the number of steps, C is the total number of class and $y_t$ is the one hot vector representation of the label at t-th epoch slice and $y_{t,c}$ is its element in class c.

## 5.2 Gated Recurrent Units

The mathematical formulation of GRU at each time step is defined as follows[5]:

$$\begin{aligned} z^{(t)} &= \sigma(W^{(z)}x^{(t)} + U^{(z)}h^{(t-1)}) \\ r^{(t)} &= \sigma(W^{(r)}x^{(t)} + U^{(r)}h^{(t-1)}) \\ h^{(t)} &= \tanh(r^{(t)} \circ Uh^{(t-1)} + Wx^{(t)}) \\ h^{(t)} &= (1-z^{(t)}) \circ h^{(t)} + z^{(t)} \circ h^{(t-1)} \end{aligned} \tag{4}$$

Where $x^{(t)}$ is the word vector embedding for input word at step t, $z^{(t)}$ is the update gate which determines the combination of new memory and previous memory carries on to next layer, $r^{(t)}$ is the reset gate which determines the proportion of new word and previous contextual information in generating new memory, $h^{(t)}$ is the new memory generated and $h^{(t)}$ is the hidden layer at step t.

Since GRU has update gate to determine the importance of new memory for current state, its prediction result is less likely to be influenced by frequent word(ideally, $z^{(t)}$=1 on frequent words without much sentiment information such as stop words ). So we output prediction at each step and use the summation of cross-entropy error at each step as loss function.

# 6  Experiments & Results

To address the skewedness problem, two different resampling methods are implemented to balance the dataset. We evaluate both resampling methods. We implement the simple RNN, the modified RNN and a GRU with Python Tensorflow and measure the train, validation, and test accuracies of each classifier we build. We visualize the hidden layer weights to see how the hidden units behaves and tune hyper parameters to improve accuracies.

## 6.1  Data Pre-processing
### 6.1.1  Sampling method 1: remove all data from the last class

Since the main source of data skewedness is the highest score class which has around ten times as many reviews as each of the rest of the classes, we employ a simple method to avoid the problem. We discard the data from the highest score class and redefine our task to predict the review score into one of the first 4 classes. The new dataset consisting of scores 1 to 4 is less biased towards higher scores.

### 6.1.2  Sampling method 2: resample data from the 4- and 5-score class

A natural way to generate a balanced dataset is to randomly sample reviews from the skewed dataset, in which case we should sample data from the 4-score and 5-score classes. According to figure 1, we would like to obtain around 4,000 reviews for each class, so we generate 4,000 random samples from the two high score classes. Now we have a more balanced dataset.

## 6.2  Implementation of RNN

Word vectors are initialized as random values uniformly distributed between [-1, 1]. The number of steps is set as 8 as recommended in the course lecture. To distinguish between different reviews, <EOS> is added at the end of each review. Then to ensure phrases of 8 words are from the same review within each epoch slice, we zero-pad the reviews to 88 words at the front of each review. Zero-padding is done at the beginning because if zero padding at the end, backpropagation will come across several identical hidden layers before propagating to an actual word, thus cause more severe vanishing gradient problem.

$L, W^{(hh)}, W^{(hx)}, b_1, W^{(s)}$ and $b_2$ are updated through the training process and applied in validation and testing. $L$ is the embedding matrix for words.

The final predicted class for each review is the class with the max value in the elements of $\hat{y}_c$, where $\hat{y}_c$ is the output prediction at the end of the corresponding review (identified by EOS).

## 6.3  Implementation of GRU
For GRU, we use the same dataset, number of steps and initialization strategy of word vectors as RNN. The training is performed on dataset with/out zero-padding.

$L$, $W^{(z)}, W^{(r)}, U^{(z)}, U^{(r)}, U$ and $W$ are updated through the training process and applied in validation and testing. $L$ is the embedding matrix for words.

The output prediction at the end of each review is used as final prediction of each class, just like RNN, to provide a fair comparison of performance.

## 6.4 Hyper-Parameters Tuning

In order to tune and find the right hyper-parameters for our model, we divide our data into three sets: a training set, a validation set for cross validation and a test set that will be used as our final prediction scores. In this section, we describe how we performed our tuning and record the accuracies depending on it. For each of the models, learning rate, L2 regularization weight and dropout value are to be tuned. Due to time and computation resource constraints, we did not tune some parameters like hidden layer size and we were not able to iteratively optimize the parameters that would have resulted in the optimal setting. Instead, we fix some parameters to reasonable values and tune the others. The following figures show the tuning results.

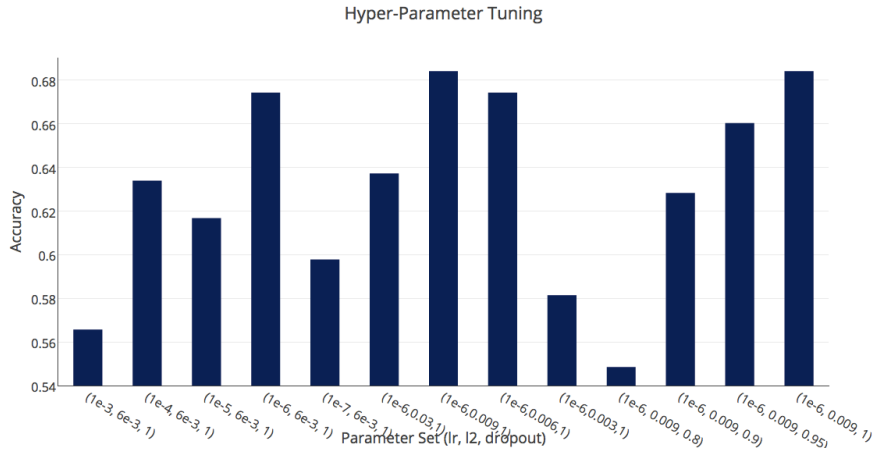

Figure 2(a). RNN(4 classes) Hyper-Parameter Tuning

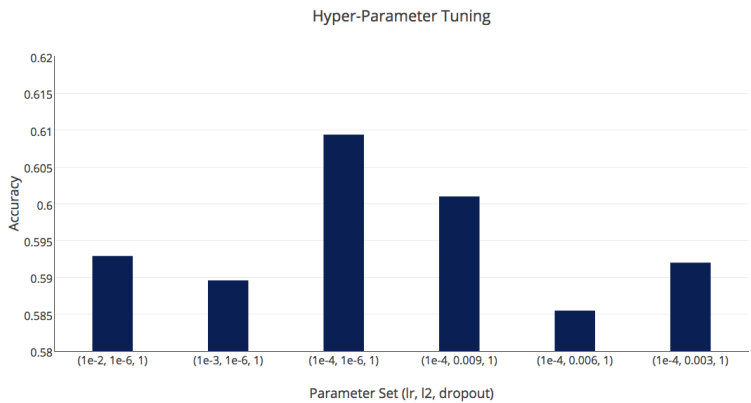

Figure 2(b). GRU(4 classes) Hyper-Parameter Tuning

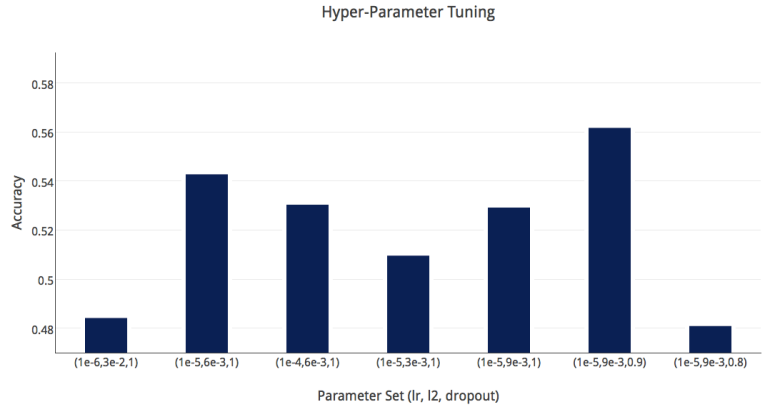

Figure 3(a). RNN(5 classes) Hyper-Parameter Tuning

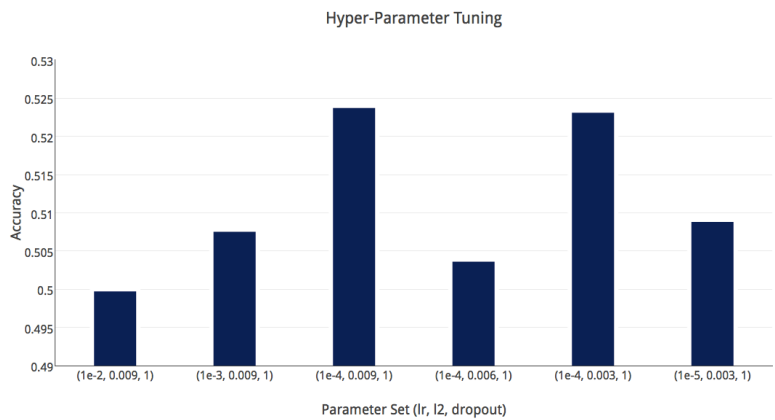

Figure 3(b). GRU(5 classes) Hyper-Parameter Tuning

The optimal set of parameters we have found for our models are as follows: RNN,4 classes(lr=$10^{-6}$, l2=0.009, dropout=1.0), RNN,5 classes(lr=$10^{-5}$, l2=0.009, dropout=0.9), GRU,4 classes(lr= $10^{-4}$, l2= $10^{-6}$, dropout=1.0), GRU,5 classes(lr= $10^{-4}$, l2=0.009, dropout=1.0). With these parameters obtained, we re-train our models and test the models. The test performances are shown in next section.

### 6.5    Accuracies

After tuning the hyper parameters, we use the optimal set of hyper parameters to train and test our model and evaluate the performance by accuracy. Accuracy is calculated by the number of correctly labeled reviews over the total number of reviews, where the predicted label at the end of a review is regarded as the final predicted label for that review. For our specific data, we have not found work on the same problem, so we don't have the state-of-the-art result. For comparison purposes, we also train RNN models with output at each step, and GRU models without zero padding.

| Model+Strategy | Training Accuracy | Test Accuracy |
|---|---|---|
| RNN(4 classes) | **93.35%** | **68.75%** |
| GRU(4 classes) | 84.31% | 61.20% |
| RNN(4 classes, w/ output at each step) | 82.14% | 60.85% |
| GRU(4 classes, w/o zero-padding) | 54.40% | 42.70% |
| **RNN(5 classes)** | **80.38%** | **51.74%** |
| GRU(5 classes) | 80.14% | 50.09% |
| RNN(5 classes, w/ output at each step) | 86.57% | 47.22% |
| GRU(5 classes, w/o zero-padding) | 43.60% | 35.70% |

Table 1. Accuracies of different models

In the 4-class prediction task, the best model in our experiment is the modified RNN. Our slightly modified RNN greatly outperforms the original RNN which outputs at each step. In the 5-class prediction task, RNN and GRU achieve comparable accuracies, whereas RNN performs slightly better than GRU. We analyzed why GRUs did not outperform modified RNN. In brief, one possible reason for this might be the lack of tuning on some of the other important hyper-parameters, like the hidden layer size and the number of steps. Another possible reason is that the reset gate might not have reset the frequent words as wanted.

### 6.6  Visualization of Hidden Layer Weights

To demonstrate the effect of training under different strategies, we present the visualization of a hidden layer at the first and last epoch in this section.

For our modified RNN, the hidden layers for different classes looks quite similar at epoch 0(shown in figure 4(a)) since the word vectors are randomly initialized. But by at the last epoch of training, the hidden layers under different labels are quite different. For instance, hidden layers under 3 and 4 star reviews have higher values around 40th dimension than hidden layers under 1 and 2 star.

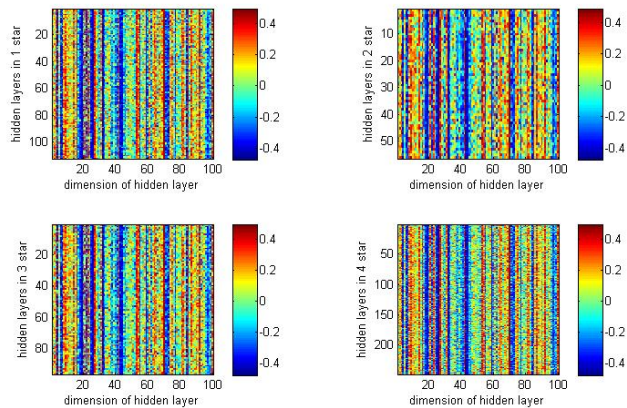

Figure 4(a). Hidden Layer under RNN at Epoch 0

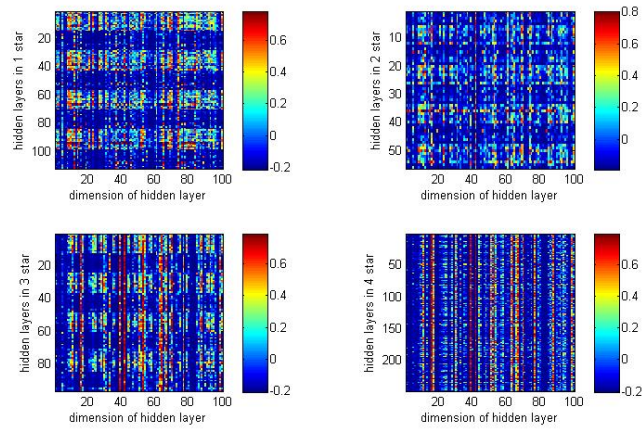

Figure 4(b). Hidden Layer under RNN at Epoch 6

For GRU, the hidden layer showed some change over the epochs, but the pattern is not as obvious as RNN, indicating a lower performance.

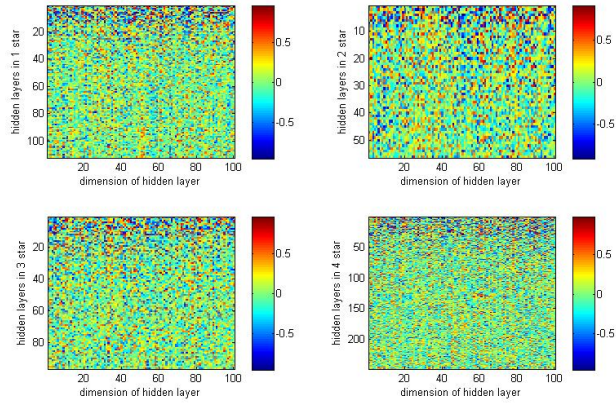

Figure 5(a). Hidden Layer under GRU at Epoch 0

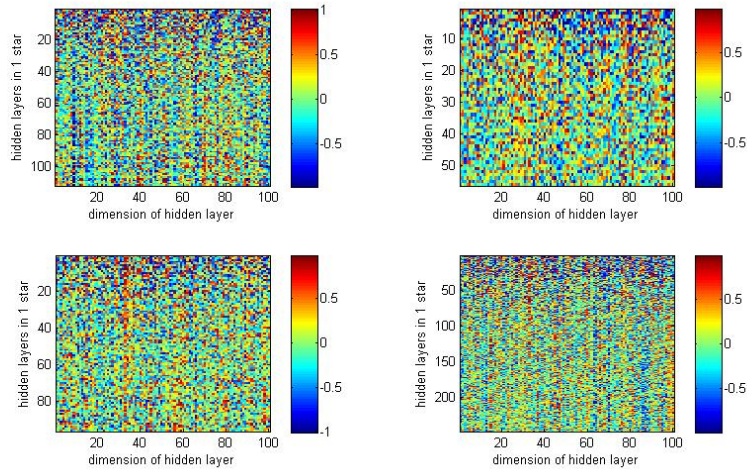

Figure 5(b). Hidden Layer under GRU at Epoch 6

# 7   Conclusion

In this paper, we present different neural network approaches including 2 versions of RNN and GRU for sentiment classification on Amazon Fine Food Reviews dataset and reach 68.75% test accuracy in 4 class classification task and 51.74% in 5 class classification task on the test set. In our experiment, we find that padding zeroes to reviews proves to be useful and the zero-padded approaches outperform the approaches without zero-padding we implement. Future work might focus on trying out more RNN models, like the bidirectional RNN and tuning other parameters like hidden layer size and number of steps.